\documentclass[10pt,twocolumn,letterpaper]{article}

\usepackage{iccv}
\usepackage{times}
\usepackage{epsfig}
\usepackage{graphicx}
\usepackage{graphics}
\usepackage{amsmath}
\usepackage{amssymb}
\usepackage{multirow}

\usepackage[ruled,vlined]{algorithm2e}

\usepackage{adjustbox}
\usepackage{booktabs}
\usepackage{subcaption}
\usepackage{tabularx}

\usepackage[breaklinks=true,bookmarks=false]{hyperref}

\iccvfinalcopy 

\def\iccvPaperID{9364} 

\newcommand*{\firstranking}{$\mathbf{1^{st}}$ }
\newcommand*{\method}{GP-S3Net}
\newcommand*{\kybernet}{$\mathbf{(AF)^2}$-S3Net}
\ificcvfinal\fi

\begin{document}

\title{GP-S3Net: Graph-based Panoptic Sparse Semantic Segmentation Network}

\author{
Ryan Razani \textsuperscript{$\ast$},
Ran Cheng  \thanks{ Indicates equal contribution.} ,
Enxu Li,
Ehsan Taghavi,
Yuan Ren,
and Liu Bingbing \thanks{
  {\tt\small Emails: \{ryan.razani, ran.cheng1, thomas.enxu.li, ehsan.taghavi, yuan.ren3, liu.bingbing\}@huawei.com}
  }\\
  Huawei Noah's Ark Lab, Toronto, Canada
}

\maketitle
\ificcvfinal\thispagestyle{empty}\fi

\begin{abstract}
   Panoptic segmentation as an integrated task of both static environmental understanding and dynamic object identification, has recently begun to receive broad research interest. 
   In this paper, we propose a new computationally efficient LiDAR based panoptic segmentation framework, called \method. GP-S3Net is a proposal-free approach in which no object proposals are needed to identify the objects in contrast to conventional two-stage panoptic systems, where a detection network is incorporated for capturing instance information. 
   Our new design consists of a novel instance-level network to process the semantic results by constructing a graph convolutional network to identify objects (foreground), which later on are fused with the background classes. Through the fine-grained clusters of the foreground objects from the semantic segmentation backbone, over-segmentation priors are generated and subsequently processed by 3D sparse convolution to embed each cluster. Each cluster is treated as a node in the graph and its corresponding embedding is used as its node feature. Then a GCNN predicts whether edges exist between each cluster pair. We utilize the instance label to generate ground truth edge labels for each constructed graph in order to supervise the learning. Extensive experiments demonstrate that GP-S3Net outperforms the current state-of-the-art approaches, by a significant margin across available datasets such as,  nuScenes and SemanticPOSS, ranking \firstranking on the competitive public SemanticKITTI leaderboard upon publication.

\end{abstract}

\section{Introduction}
\label{sec:intro}
One of the main tasks in realizing autonomy in robotic applications is scene understanding using the available data collected from sensors such as camera, Light Detection And Ranging (LiDAR) and RAdio Detection and Ranging (RADAR). Scene understanding can be divided into different tasks such as scene classification, object detection, semantic/instance segmentation and more recently panoptic segmentation. In recent years, deep learning has been used as the main solution to object detection and segmentation. With more granular data available, specially from high resolution LiDARs, the task of semantic/instance segmentation has become a more vital part of any perception module due to generating all the fundamental perceptual information needed for robotic applications such as free space, parking area and vegetation, in addition to all dynamic objects in 3D coordinates.

\begin{figure}[htbp!]
    \centering 
    \includegraphics[width=\linewidth]{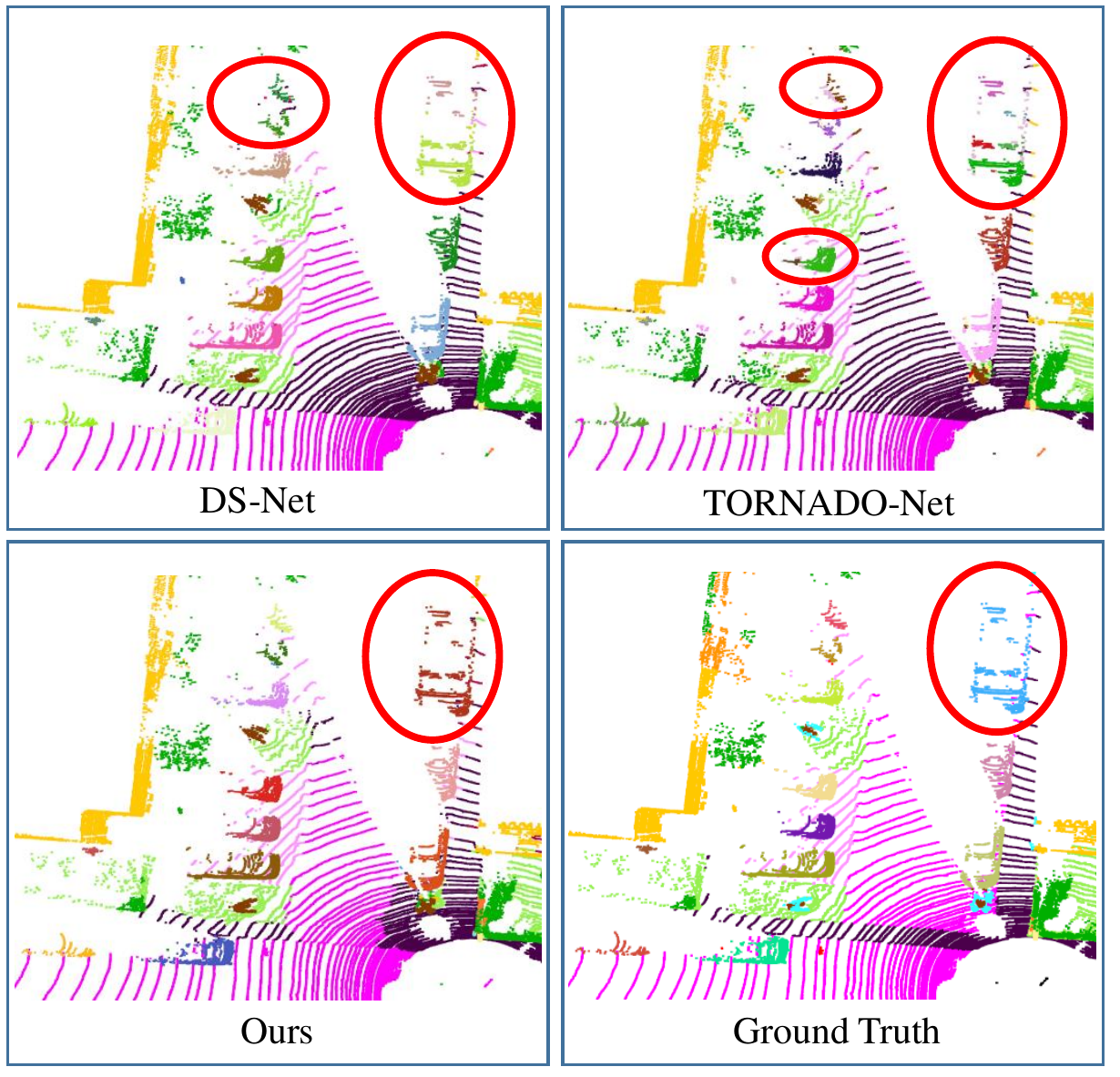} 
    \caption{Comparison of our proposed method with DS-Net \cite{hong2020lidar} and TORNADONet \cite{gerdzhev2020tornadonet} on SemanticKITTI benchmark \cite{DBLP:conf/iccv/BehleyGMQBSG19}.
      DS-Net and TORNADONet suffer from over-segmentation problem, while ours shows successful panoptic predictions.
      }
    \label{fig:kyber_intro}
\end{figure}

Normally, the task of semantic segmentation and instance segmentation are treated separately, using two different DNN models. In contrast, panoptic segmentation combines semantic and instance segmentation tasks as explained in \cite{kirillov2019panoptic} by defining two different categories of model predictions, namely, ``things'' and ``stuff''. ``Things'' is referred to all countable objects such as cars, pedestrians, bikes, etc. and ``stuff'' is referred to uncountable semantics (background) such as road, vegetation, parking. Then, the panoptic task is defined by predicting the things and stuff together in one unified solution. Although the panoptic segmentation definition seems plausible, the question of ``Is there a way to integrate independent semantic segmentation and target recognition into a system so that the instance level segmentation of objects can effectively use the results of semantic segmentation?" has not been addressed adequately in the literature.

In this paper we present a novel approach to solve the panoptic segmentation problem solely based on LiDAR that can achieve state-of-the-art panoptic and semantic segmentation results on various benchmarks such as SemanticKITTI \cite{DBLP:conf/iccv/BehleyGMQBSG19}, nuScenes \cite{caesar2020nuscenes}, and SemanticPOSS \cite{pan2020semanticposs}. 
Our method uses graph network to generate instances directly from
over-segmented clusters obtained by HDBSCAN \cite{campello_hdbscan} that takes
in predicted foreground points from the semantic segmentation backbone. Instead of learning the offset vector of each cluster and obtaining instances by re-clustering, a connectivity probability of each edge is learned in the graph. Over-segmented
clusters are aggregated together by these connected edges
to form instances. Thus, conventional clustering problem is
transformed into a supervised edge classification problem.
By doing so, we introduce the following contributions:
\begin{itemize}
    \item A flexible panoptic segmentation framework to benefit from the best available semantic segmentation models and their output.
    \item A novel clustering and graph convolutional neural network that generates instance-level results (things and their IDs).
    \item A seamless fusion of semantic and instance level results to generate panoptic segmentation predictions.
    \item A thorough experimental results of three major outdoor datasets and an ablation study to show the effectiveness of the proposed solution.
\end{itemize}

\section{Related work}
\label{sec:related}

\noindent \textbf{Semantic segmentation}
Most point cloud semantic segmentation methods can be categorized to projection-based, point-based and voxel-based methods, by their method of data processing. Projection-based methods aim to project 3D point clouds into 2D image space in either the top-down Bird-Eye-View \cite{8403277, 10.1007/978-3-030-11009-3_11}, or spherical Range-View (RV) \cite{cortinhal2020salsanext, milioto2019rangenet++, razani2021litehdseg} or multi-view format \cite{gerdzhev2020tornadonet}, and then process with the standard 2D CNNs. Benefited from the quick arrangement of unordered points and the speed of 2D CNN, it is easy for projection-based methods to achieve real time performance, although their accuracy is limited by the information loss during projection. Point-wise methods, such as PointNet++\cite{qi2017pointnet++}, KPConv \cite{thomas2019kpconv} and RandLA-Net \cite{hu2020randla}, process the raw 3D points without applying any additional transformation or pre-processing. This kind of methods are more suitable for small scale point clouds. However, for large scale point clouds, due to large computation and memory  requirements, it is usually difficult to make real time inference. Voxel-based approaches transform a point cloud into 3D volumetric grids in which 3D convolutions are used. Theoretically, 3D convolution is a natural extension of the 2D convolution concept. 3D volumetric grids embedded in Euclidean space can guarantee the shift invariant (or space invariant). However, the high computational requirements still constrain development of the voxel-based approaches. Recently, several sparse convolution libraries, such as Torch sparse, Minkowski Engine \cite{choy20194d} and SPConv, have been developed to accelerate the convolution computation by fully exploiting sparsity of point clouds. With the help of these libraries, voxel-based methods such as \kybernet~ \cite{cheng2021af2s3net}, Cylinder3D \cite{zhou2020cylinder3d} and S3Net \cite{cheng2021s3net} have achieved state-of-the-arts LiDAR semantic segmentation performance on public datasets. 

\noindent \textbf{Instance segmentation}
Instance segmentation of point clouds can be divided into two categories, i.e. proposal-based and proposal-free methods. Proposal-based methods, e.g. 3D Bonet \cite{NEURIPS2019_d0aa518d} and  GSPN \cite{YiLi_GSPN_2019}, implement 3D bounding box detection followed by a refinement in the bounding box. The refinement network generates a point-wise mask, which filters out the points not belong to the instance. Proposal-free methods are based on clustering. The whole scene is usually segmented into small blocks, and the embeddings of each block are learned and used in the final clustering. PointGroup \cite{Jiang_2020_CVPR} clusters points towards the instance centroid by learning the offsets of each block. OccuSeg \cite{9157103} divides the whole scene into super-voxels and uses a graph-based method for clustering.

\noindent \textbf{Panoptic segmentation}
Panoptic segmentation unifies both semantic segmentation and instance segmentation. Panoptic segmentation can also be divided into proposal-based (top-down) and proposal-free (bottom-up) methods. Proposal-based methods such as EfficientLPS \cite{sirohi2021efficientlps} and MOPT \cite{hurtado2020mopt}, are two-stage approaches. They detect the foreground instances first, then refine these instances and finally implement the semantic segmentation on the background stuff. In contrast, proposal-free methods are usually single-staged. They segment the scene semantically and identify the instances based on the semantic segmentation result by using a unified network. This kind of network usually consists of a single backbone for feature extraction and two separate branches for semantic and instance segmentation tasks, which are also known as semantic head and instance head. Panoster \cite{gasperini2021panoster}, LPSAD \cite{9340837} and DS-Net \cite{hong2020lidar} use this two separate branches structure. Among them, Panoster can use KPConv \cite{thomas2019kpconv} and SalsaNext \cite{cortinhal2020salsanext} as its backbone. All these three methods achieve competitive performance on outdoor LiDAR datasets. Recently, some algorithms use graph to represent the high level topological relationship between foreground things and background stuff, e.g. Wu \textit{et al.} \cite{Wu_2020_CVPR} proposed a bidirectional graph connection module to diffuse information across branches, and achieved state-of-the-art performance on the COCO and ADE20K panoptic segmentation benchmarks.

\section{Proposed approach}
\label{sec:method}

In this section, we first introduce the problem formulation. Next, the proposed method, \method, is presented with detailed description of its components along with network optimization details.

\subsection{Problem statement}
\label{sec:problem}
Here we consider a panoptic segmentation task in which a LiDAR point cloud frame is given with a set of unordered points $(P, L) = (\{ p_{i},  l_{i} \})$ with $p_{i} \in \mathbb{R}^{d_{in}}$ and $i = 1, ... , N$, where $N$ denotes the number of points in a point cloud scan. Each point can be paired with features such as Cartesian coordinates $(x,y,z)$, intensity of returning laser beam $(i)$, colors $(R,G,B)$, etc.  Here, $l_{i} \in  \mathbb{R}^2$ represents the ground truth labels corresponding to each point $p_{i}$ which can be either ``stuff'' with a single class label $\mathcal{L}$, or ``things'' with a single class label $\mathcal{L}$ and an instance label $\mathcal{J}$.

Our goal is to learn a function $\mathcal{F}_{cls,inst}(., \Phi)$ parameterized by $\Phi$ that assigns a single class label $\mathcal{L}$ for all the points related to ``stuff'' and, class label $\mathcal{L}$ and an instance label $\mathcal{J}$ for all the points in the point cloud related to ``things''. In other words, $\mathcal{F}_{panoptic}(., \Phi)$ assigns class and instance label pair $(\hat{c}_{i},\hat{inst}_{i})$ to each point $p_{i}$ where applicable. To solve this problem, we present \method~ in which a proposal-free design is leveraged to generate predictions for panoptic segmentation benefiting from a novel GNN approach.

\label{sec:arch}
\begin{figure}[h]
    \centering
    \includegraphics[width=\linewidth]{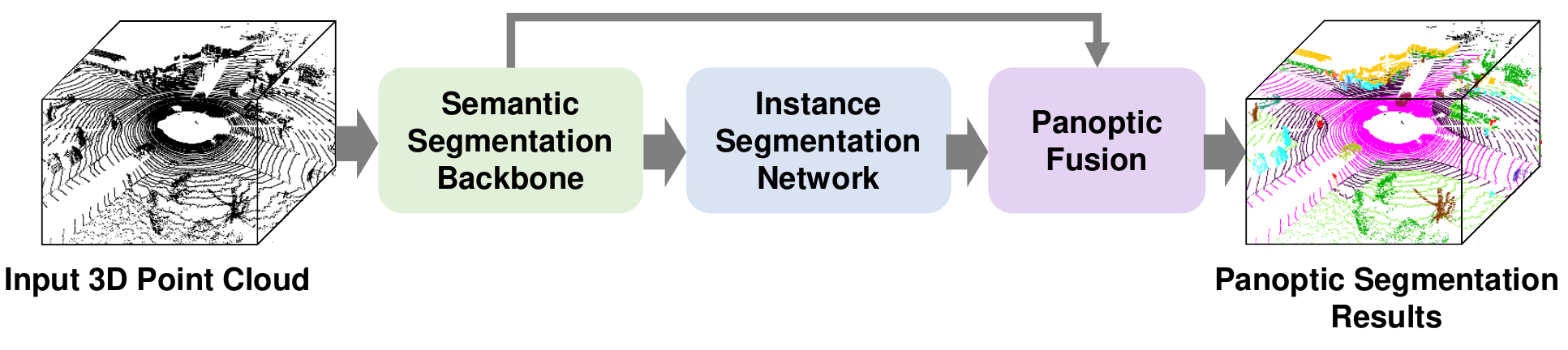}
    \caption{Overview of our proposed \method~ model for LiDAR panoptic segmentation}.
    \label{fig:GPNet}
\end{figure}

\begin{figure*}
    \centering
    \includegraphics[width=\linewidth]{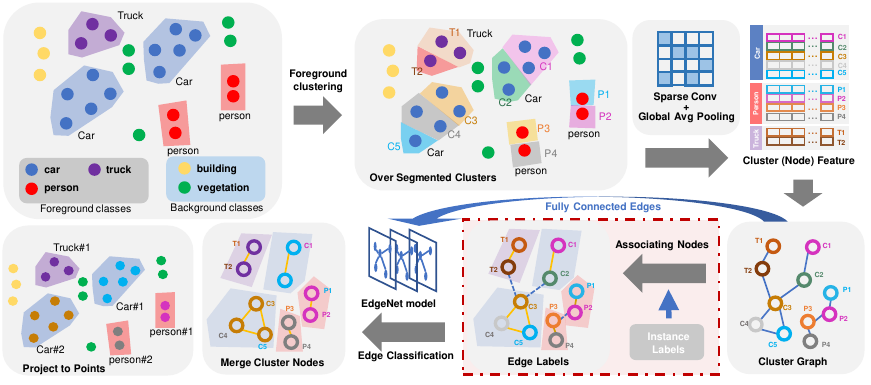}
    \caption{Illustration of \method~instance learning procedure. 
    The input to the instance network is semantically segmented point cloud from the semantic network. The output is the list of instance clusters for foreground classes which will be fused with background classes to form panoptic results. Instance labels generation block drawn in red dashed box is ON only during training.
    }
    \label{fig:overall_model}
\end{figure*}

\subsection{Network architecture}

The panoptic segmentation of 3D point cloud usually simultaneously performs semantic segmentation and object recognition on the point cloud, and uses the result of object recognition as a mask to classify the instance of foreground semantic categories (such as vehicles, pedestrians). We propose a cluster method based on graph neural network supervised learning to achieve adaptive object instance segmentation. Through the fine-grained cluster of the foreground class, the over segmentation prior is obtained, and then the three-dimensional sparse convolution is used to embed each cluster. This embedding is used as a feature of the graph node to establish an edge between multiple clusters in the same instance. Essentially, we use semantic information to establish a graph structure. Since the instance label is known, we can perform supervised learning on edges to predict the existence of edges between two cluster embeddings. In essence, we convert an unsupervised task into a supervised learning task. This method is applicable to all panoptic segmentation, including point cloud, rgb image.

The overall pipeline of our proposed \method~ is shown in Figure~\ref{fig:GPNet}.
It includes a semantic segmentation segmentation backbone followed by a graph-based instance segmentation network and panoptic fusion. A generic semantic LiDAR segmentation module takes in raw LiDAR point cloud as input and outputs semantic masks which are directly used in the new enhanced instance model. The instance model then clusters the ``things'' semantic masks and encodes their features into a unified feature vector. These features are fed into a GCNN model followed by a merging step to create the instance information. Finally, the output of the semantic and instance segmentation network are fused (concatenated) without including any learned-based model. The final output is panoptically labeled point cloud in which every point has both a semantic and a instance label. The detailed description of the network architecture is provided below.

\subsubsection{Semantic segmentation backbone}
For the semantic segmentation model we benefit from the model introduced in \cite{cheng2021af2s3net}.
\kybernet~ consists of an end-to-end 3D encoder-decoder CNN network that combines the voxel-based and point-based learning methods into a unified framework, resulting in highly accurate semantic segmentation model with competitive results for various datasets such as SemanticKITTI and nuScenes.

\kybernet~ provides semantic segmentation results using efficient feature attention and fusion at different scales, making the model able to predict semantics for all classes, especially smaller dynamic objects on the road. This characteristic of \kybernet~ makes it a viable choice for the first part of the proposed panoptic framework introduced in this paper.

\subsubsection{Instance segmentation network}

\begin{figure}
    \centering
    \includegraphics[width=\linewidth]{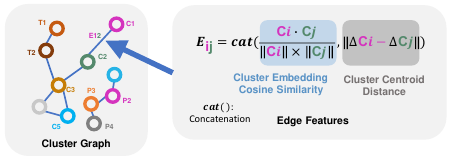}
    \caption{Computation of the edge features used for constructing a cluster graph. $C_i$ and $C_j$ are the feature embedding of node $i$ and node $j$ for a given edge $E_{ij}$. $\Delta C$ is the 3d coordinates of point cloud centroids of each node.} 
    \label{fig:edge_features}
    \vspace{-5px}
\end{figure}

As shown in the top branch of Figure~\ref{fig:overall_model}, our method first uses the segmented point cloud as input and performs over-segmentation by clustering on foreground classes (e.g., car, pedestrian). Then, the over-segmented clusters are fed into a sparse convolutional neural network followed by a global average pooling layer to embed the cluster points and their corresponding features. Each point of the cluster has point-wise normal features, intensity and inherited semantic probabilities. All points in one cluster will be converted into a $1 \times K$ embedding vector. Since we use global average pooling, the cluster embedding can compress any number of points into a single point embedding vector of features. We treat the clusters as graph nodes and build a fully connected graph out of all clusters. The embedding vector is thus regarded as the graph node feature. The edge features are depicted in Figure~\ref{fig:edge_features}, where the edge feature is defined by the cosine similarity of two cluster embedding vectors and the Euclidean distance of two clusters' centroids.

\begin{figure}
    \centering
    \includegraphics[width=\linewidth]{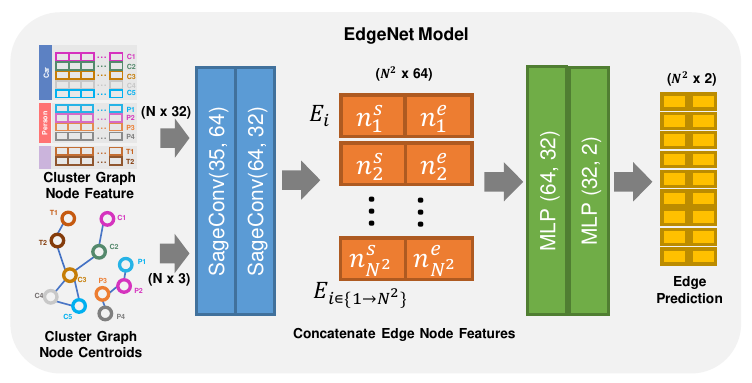}
    \caption{EdgeNet model}
    \label{fig:edgenet_model}
    \vspace{-5px}
\end{figure}

Instance segmentation can be addressed by clustering the objects' semantic labels, however the clustering process is not supervised friendly in which only a clustering radius can be tuned. Therefore, it is difficult to exploit the instance labels offered in most  of the datasets.
In our method, we convert this unsupervised clustering task into a supervised edge classification problem. First, the foreground class points are passed though HDBSCAN
so that each instance is divided into several over-segmented clusters. Then, the clustered points, which belong to the same instance object, are associated using the instance ground truth labels, as illustrated in Algorithm~\ref{algo:cluster_node_association}. Since we have built the fully connected graph, as illustrated in the bottom branch of Figure~\ref{fig:overall_model}, we can create the ground truth labels of edges using the associating relationship between cluster nodes. The edges between the associated clusters are thus marked as 1, otherwise 0. Finally after getting the ground truth labels and features of graph nodes and edges, we apply the edge classification using a graph convolutional neural network, named EdgeNet, to predict fully connectected edge labels. After the predition, those  over-segmented clusters are merged into a corresponding instance node and the instance ID is projected to each points inside the nodes as output.    

Instance-level segmented regions are geometrically mutual exclusive, thus we can leverage the connectivity of sub-graph to describe the clusters that belongs to a single instance. Therefore, the local connectivity can be described by a binary adjacent matrix where $1$ means connection exists and two corresponding nodes belongs to the same instance object. As a result, we can recover the instance points from merging the connected cluster nodes to the same instance group.
In this manner, we predict the edges of a fully connected cluster graph in a supervised learning fashion. As depicted in Figure~\ref{fig:edgenet_model}, EdgeNet model takes the fully connected cluster graph as input and predicts the edge class for each two pairs of cluster graph. We concatenate the cluster node embedding features ($N \times 32$), where $N$ is the number of clusters predicted by HDBSCAN, and the node centroid feature ($N \times 3$) as the input graph node feature ($N \times 35$) of the EdgeNet model. The edge feature has two components, the cosine similarity of the cluster graph node feature and the euclidean distance of the cluster graph node centroids. 
The fully connected cluster graph with node and edge features are fed into two SageConv layers \cite{hamilton2017inductive} to perform neighborhood node feature aggregation. Then, the starting node feature $n_{i}^{s}$ and ending node feature $n_{i}^{e}$ of a given edge $E_{i}$, where $i \in \{1,2,..., N^{2} \},$ are concatenated as edge feature of the edge $E_{i}$ with size of ($1 \times 64$). Finally, the edge features are passed into a fully connected layer and the binary edge class labels are predicted. The details of the EdgeNet GCNN architecture is presented in Table \ref{tab:EdgeNet_design}.

\begin{table}[htb]
\begin{center} 
\scalebox{0.45}
{
\begin{tabular}{|l|l|l|l|l|l|l|}
\hline
Stages               & Operation Layer            & Operation Layer           & ($F_{in}, F_{out}$) & Filter Size & Stride & Output Size      \\ \hline \hline
Input                & Input Foreground Point Cloud  & -                           &  -                  &  -                    &  -             & Mx24                      \\ \hline
SE     & SparseConv layer           & Sparse Convolution + ReLU & 24,64             & 3x3x3               & 1x1x1        & Mx64                      \\ \hline
SE     & SparseConv layer           & Sparse Convolution + ReLU & 64,32             & 3x3x3               & 1x1x1        & Mx32                     \\ \hline
SE     & Sparse Global Pooling       & Average pooling           & -                 & cluster-wise               & -        & Nx32                      \\ \hline
GNN & Graph Conv Mean Aggregator & SageConv + ReLU           & 35,64             & -                   & -            & Nx64  \\ \hline
GNN & Graph Conv Mean Aggregator & SageConv                  & 64,32                & -                   & -            & Nx32                      \\ \hline
GNN & EdgeNet                    & MLP + ReLU                & (32+32),32        & -                   & -            & (NxN)x32                  \\ \hline
GNN & EdgeNet                    & MLP + Softmax             & 32,2              & -                   & -            & (NxN)x2                   \\ \hline
output               & -                          & -                         & -                 & -                   & -            & (NxN)x2                   \\ \hline
\end{tabular}
}
\end{center}
\caption{EdgeNet Design. We denote SE as Sparse Embedding, GNN as Graph Neural Network, M and N are number of foreground points and number of cluster, respectively. ($F_{in}, F_{out}$) are the number of input and output features. }
\label{tab:EdgeNet_design}
\vspace{-5px}
\end{table}

As our method relies on the oversegmentation process, it might be possible that the predictions from the semantic segmentation backbone are incorrect when the GNN thinks multiple clusters with different semantics belong to a single instance. Thus, we refine the semantics of each instance with point-wise majority voting. 

\textbf{Graph edge labels generation} The graph nodes are obtained by running HDBSCAN on the foreground point cloud. Here, we assume the points within the same cluster (graph node) belong to the same instance. In order to supervise the edge classification of the graph, we construct the binary ground truth (GT) edge label matrix, $l_{edge} \in \mathbb{B}^{N \times N}$ where a \textit{True} entry indicates the two nodes belong to the same object, \textit{False} indicates they do not, and $N$ is the number of nodes. Note that this GT edge label is only needed during training, and is obtained by associating each graph node with the GT instance label. Detailed pseudo-code of this operation can be found in Algorithm \ref{algo:cluster_node_association}. In summary, we build a correspondence between the cluster ID of the point and its GT instance ID using the \textit{offset} trick. In particular, the most 32 bit is the GT ID and the least 32 bit is the cluster ID for each entry in $s_{combo}$. Thus, we can obtain the list of nodes belong to the same object, and a permutation of each node pair in the list will be marked as positive labels.

\begin{algorithm}
\SetAlgoLined
\textbf{Input:} \textit{Things} Points $p \in \mathbb{R}^{M \times 3}$,
\textit{Things} Instance Label $l \in \mathbb{N}^{M} $\;
\textbf{Output:} Binary labels $l_{edge} \in \mathbb{B}^{N \times N}$ of fully connected edge between each cluster\;

 $c := hdbscan(p)$\;
 $nCluster := len(set(c))$\;
 $l_{edge} := zeros(nCluster, nCluster)$\;
 $d := \{\}$\;
 \For{$i'_{gt}$ \textbf{in} $set(l)$}{
$d[i'_{gt}]:= \{\}$
}
 $\text{offset} := 2^{32}$\;

$s_{combo} := c + \text{offset} * l$\;
$s' := set(s_{combo})$\;
$s'_{gt} := s' // \text{offset} $\;
$s'_{pred} := s' \% \text{offset} $\;
\For{$idx$, $i'_{gt}$ \textbf{in} $enum(s'_{gt})$}{
$d[i'_{gt}].append(s'_{pred}[idx]$)\;
}

\For{$i'_{gt}$ \textbf{in} $d$}{
    $trueEdge := permute(d[i'_{gt}])$\;
    \For{$(i, j)$ \textbf{in} $trueEdge$}{
        $l_{edge}[i,j] = True$\;
    }
}

 \caption{Cluster node association}
 \label{algo:cluster_node_association}
\end{algorithm}

\section{Experimental Results}
\label{sec:experiments}
In this section, we firstly introduce the three datasets that we report the results on, followed by the quantitative metrics we use. Secondly, we present our experimental setup in detail. Further, the quantitative and qualitative results together with ablation analysis on various clustering algorithms as baselines are shown.

\textbf{SemanticKITTI} is the first available benchmark on LiDAR-based panoptic segmentation \cite{DBLP:conf/iccv/BehleyGMQBSG19}. It contains 22 sequences where sequence 00-10 are used for training and validation (19,130 training frames and 4,071 validation frames) and 11-21 are held out for testing (20,351 frames). We validate and provide ablation analysis on sequence 08. Each point in the dataset is provided with a semantic label of 28 classes which are mapped to 19 classes for the task of panoptic segmentation. Among these 19 classes, 11 of them are \textit{stuff} classes and 8 of them are \textit{things} where the instance IDs are available.

\textbf{nuScenes} is a multi-modal autonomous driving dataset containing 1000 scenes where 700 are used for training, 150 for validation, and the rest for testing \cite{caesar2020nuscenes}. Since the dataset does not provide point-level panoptic labels for LiDAR scans, we obtain the panoptic segmentation labels by the following procedure. With semantic labels from the \textit{lidarseg} dataset and 3D bounding box annotations from the \textit{detection} dataset, we assign every point having the same semantic inside each bounding box unique instance IDs. Due to the test set labels being held out, we train the model on the training split (28,130 frames) and evaluate on the validation split (6,019 frames). Among 16 labeled classes in \textit{lidarseg}, 8 vehicle and human classes are \textit{things} and the rest are considered \textit{stuff}.

\textbf{SemanticPOSS} is a dataset with data collected in Peking University and uses the same data format as SemanticKITTI. It contains 2,988 LiDAR scans (six sequences) with point-wise semantic label and instance ID. Sequence 00, 01, 03, 04 and 05 are used for training, and sequence 02 is the validation set. 11 remapped classes are used in panoptic segmentation, among which 3 classes are \textit{things}, and 8 classes are \textit{stuff}. Comparing with SemanticKITTI and nuScenes, SemanticPOSS has a large quantity of dynamic instances \cite{pan2020semanticposs} thus challenging for panoptic segmentation tasks. 

\textbf{Evaluation metric} Mean Panoptic Quality (PQ) is used as our main metric to evaluate and compare the results with others. PQ can be further decomposed into SQ and RQ to provide additional insights of the results as shown in Equation \ref{pqsqrq} \cite{panopticMetric}. These three metrics are calculated separately on \textit{stuff} and \textit{things} classes, providing PQ$^{St}$, SQ$^{St}$, RQ$^{St}$ and PQ$^{Th}$, SQ$^{Th}$, RQ$^{Th}$.

\begin{equation}
    \textbf{PQ}_{c} = 
    \underbrace{\frac{\sum_{(p,g) \in TP_{c}}^{} \textbf{IoU}(p,g)}{|TP_{c}|}}_\text{Segmentation Quality (\textbf{SQ})}  
    \times
    \underbrace{\frac{|TP_{c}|}{|TP_{c}|  + \frac{1}{2}|FP_{c}| + \frac{1}{2}|FN_{c}|}}_\text{Recognition Quality (\textbf{RQ})} 
    \label{pqsqrq}
\end{equation}

\begin{equation}
    \textbf{PQ} = \frac{1}{n}\sum_{c=1}^{n} \textbf{PQ}_{c}
    \label{pqmean}
\end{equation}

\noindent where $n$ denotes the total number of classes, $(p,g)$ represent the prediction and ground truth, and  $|TP_{c}|$, $|FP_{c}|$, $|FN_{c}|$ are the set of true positive, false positives, and false negative matches for class $c$ respectively. A match is a true positive if their IoU is larger than 0.5. In addition, PQ$^{\dagger}$ is also reported as suggested in \cite{pqDagger}.

\subsection{Experimental setup}
We train our model using SGD optimizer with momentum of $0.9$ and learning rate of $0.001$, weight decay of $0.0005$ for $150$ epochs for all three datasets. Experiments are done on NVIDIA V$100$ GPUs. Cross Entropy Loss is used to train the edge classification for the GCNN, while the semantic segmentation is supervised by losses presented in \cite{cheng2021af2s3net}. Embedding size $K$ for the graph node is set to 32.

We implement LPSAD as a baseline model and results on SemanticKITTI validation set were matched as reported in \cite{9340837}. We also provide the quantitative and qualitative results of this baseline model on nuScenes and SemanticPOSS. In addition, to generate a better panoptic method using range-image data, we train the TORNADO-Net \cite{gerdzhev2020tornadonet} (without the PPL block) for the task of panoptic segmentation as it represents a better range-image segmentation results. Further, we train DS-Net \cite{hong2020lidar} on SemanticPOSS with the official codebase released by the authors.

\begin{table*}[htbp!]
{
\centering
\resizebox{1.82\columnwidth}{!}{
\begin{tabular}{l|cccc|ccc|ccc|cccccccccc}
\hline 
Method & PQ  
& PQ\textsuperscript{$\dagger$}
& RQ
& SQ 
& PQ\textsuperscript{Th} 
& RQ\textsuperscript{Th}
& SQ\textsuperscript{Th}  
& PQ\textsuperscript{St}  
& RQ\textsuperscript{St}  
& SQ\textsuperscript{St}  
&  mIoU  \\
\hline
RangeNet++ \cite{milioto2019rangenet++} + PointPillars \cite{Lang_2019_CVPR_pointpillars}
& $37.1$ & $45.9$ & $47.0$ & $75.9$ & $20.2$ & $25.2$ & $75.2$ & $49.3$ & $62.8$ & $76.5$ & $52.4$  \\
LPSAD \cite{9340837}
& $38.0$ & $47.0$ & $48.2$ & $76.5$ & $25.6$ & $31.8$ & $76.8$ & $47.1$ & $60.1$ & $76.2$ & $50.9$  \\
PanopticTrackNet \cite{hurtado2020mopt}
& $43.1$ & $50.7$ & $53.9$ & $78.8$ & $28.6$ & $35.5$ & $80.4$ & $53.6$ & $67.3$ & $77.7$ & $52.6$ \\
KPConv \cite{thomas2019kpconv} + PointPillars \cite{Lang_2019_CVPR_pointpillars}
& $44.5$ & $52.5$ & $54.4$ & $80.0$ & $32.7$ & $38.7$ & $81.5$ & $53.1$ & $65.9$ & $79.0$ & $58.8$ \\
Panoster \cite{gasperini2021panoster}
& $52.7$ & $59.9$ & $64.1$ & $80.7$ & $49.4$ & $58.5$ & $83.3$ & $55.1$ & $68.2$ & $78.8$ & $59.9$  \\
DS-Net \cite{hong2020lidar}
& $55.9$ & $62.5$ & $66.7$ & $82.3$ & $55.1$ & $62.8$ & $87.2$ & $56.5$ & $69.5$ & $78.7$ & $61.6$  \\
EfficientLPS \cite{sirohi2021efficientlps}
& $57.4$ & $63.2$ & $68.7$ & $\textbf{83.0}$ & $53.1$ & $60.5$ & $\textbf{87.8}$ & $\textbf{60.5}$ & $\textbf{74.6}$ & $\textbf{79.5}$ & $61.4$  \\
\hline
 \textbf{\method} [\textcolor{blue}{Ours}] 
 & $\textbf{60.0}$ & $\textbf{69.0}$ & $\textbf{72.1}$ & $82.0$ & $\textbf{65.0}$ & $\textbf{74.5}$ & $86.6$ & $56.4$ & $70.4$ & $78.7$ & $\textbf{70.8}$ 
\\
\hline
\end{tabular}
}
\caption{Comparison of LiDAR panoptic segmentation performance on SemanticKITTI\cite{DBLP:conf/iccv/BehleyGMQBSG19} test dataset.}
\label{bigtable} }
\end{table*} 

\begin{table*}[htbp!]
{
\centering
\resizebox{1.82\columnwidth}{!}{
\begin{tabular}{l|cccc|ccc|ccc|cccccccccc}
\hline 
Method & PQ  
& PQ\textsuperscript{$\dagger$}
& RQ
& SQ 
& PQ\textsuperscript{Th} 
& RQ\textsuperscript{Th}
& SQ\textsuperscript{Th}  
& PQ\textsuperscript{St}  
& RQ\textsuperscript{St}  
& SQ\textsuperscript{St}  
&  mIoU  \\
\hline
RangeNet++ \cite{milioto2019rangenet++} + PointPillars \cite{Lang_2019_CVPR_pointpillars}
& $36.5$ & $-$ & $44.9$ & $73.0$ & $19.6$ & $24.9$ & $69.2$ & $47.1$ & $59.4$ & $75.8$ & $52.8$  \\
LPSAD \cite{9340837} [Our Implementation]
& $37.4$ & $44.2$ & $47.8$ & $66.9$ & $25.3$ & $32.4$ & $65.2$ & $46.2$ & $58.9$ & $68.2$ & $49.4$  \\
PanopticTrackNet \cite{hurtado2020mopt}
& $40.0$ & $-$ & $48.3$ & $73.0$ & $29.9$ & $33.6$ & $76.8$ & $47.4$ & $59.1$ & $70.3$ & $53.8$ \\
KPConv \cite{thomas2019kpconv} + PointPillars \cite{Lang_2019_CVPR_pointpillars}
& $41.1$ & $-$ & $50.3$ & $74.3$ & $28.9$ & $33.1$ & $69.8$ & $50.1$ & $62.8$ & $77.6$ & $56.6$ \\
TORNADO-Net \cite{gerdzhev2020tornadonet} w/ Fusion
& $50.6$ & $55.9$ & $62.1$ & $74.9$ & $48.1$ & $57.5$ & $72.5$ & $52.4$ & $65.4$ & $76.7$ & $59.2$  \\
Panoster \cite{gasperini2021panoster}
& $55.6$ & $-$ & $66.8$ & $79.9$ & $56.6$ & $65.8$ & $-$ & $-$ & $-$ & $-$ & $61.1$  \\

DS-Net \cite{hong2020lidar}
& $57.7$ & $63.4$ & $68.0$ & $77.6$ & $61.8$ & $68.8$ & $78.2$ & $54.8$ & $67.3$ & $77.1$ & $63.5$  \\
EfficientLPS\cite{sirohi2021efficientlps} 
& $59.2$ & $65.1$ & $69.8$ & $75.0$ & $58.0$ & $68.2$ & $78.0$ & $\textbf{60.9}$ & $71.0$ & $72.8$ & $64.9$  \\
\hline
 \textbf{\method} [\textcolor{blue}{Ours}] 
 & $\textbf{63.3}$ & $\textbf{71.5}$ & $\textbf{75.9}$ & $\textbf{81.4}$ & $\textbf{70.2}$ & $\textbf{80.1}$ & $\textbf{86.2}$ & $58.3$ & $\textbf{72.9}$ & $\textbf{77.9}$ & $\textbf{73.0}$ 
\\
\hline
\end{tabular}
}
\caption{Comparison of LiDAR panoptic segmentation performance on SemanticKITTI\cite{DBLP:conf/iccv/BehleyGMQBSG19} validation dataset.}
\label{bigtable_val}}
\end{table*} 

\begin{table*}[htbp!]
{
\centering
\resizebox{1.82\columnwidth}{!}{
\begin{tabular}{l|cccc|ccc|ccc|cccccccccc}
\hline 
Method & PQ  
& PQ\textsuperscript{$\dagger$}
& RQ
& SQ 
& PQ\textsuperscript{Th} 
& RQ\textsuperscript{Th}
& SQ\textsuperscript{Th}  
& PQ\textsuperscript{St}  
& RQ\textsuperscript{St}  
& SQ\textsuperscript{St}  
&  mIoU  \\
\hline
Cylinder3D \cite{zhou2020cylinder3d} + PointPillars \cite{Lang_2019_CVPR_pointpillars}
& $36.0$ & $44.5$ & $43.0$ & $83.3$ & $23.3$ & $27.0$ & $83.7$ & $57.2$ & $69.6$ & $82.7$ & $52.3$ \\
Cylinder3D \cite{zhou2020cylinder3d} + SECOND \cite{yan_mao_li_2018_second}
& $40.1$ & $48.4$ & $47.3$ & $84.2$ & $29.0$ & $33.6$ & $84.4$ & $58.5$ & $70.1$ & $83.7$ & $58.5$ \\
DS-Net \cite{hong2020lidar}
& $42.5$ & $51.0$ & $50.3$ & $83.6$ & $32.5$ & $38.3$ & $83.1$ & $59.2$ & $70.3$ & $\textbf{84.4}$ & $70.7$  \\
RangeNet++ \cite{milioto2019rangenet++} + Mask R-CNN \cite{he2017mask}
& $45.2$ & $53.7$ & $55.7$ & $80.2$ & $39.2$ & $48.6$ & $79.4$ & $55.2$ & $67.6$ & $81.5$ & $61.7$  \\
PanopticTrackNet \cite{hurtado2020mopt}
& $50.0$ & $57.3$ & $60.6$ & $80.9$ & $45.1$ & $52.4$ & $80.3$ & $58.3$ & $74.3$ & $81.9$ & $63.1$ \\
KPConv \cite{thomas2019kpconv} + Mask R-CNN \cite{he2017mask}
& $50.1$ & $57.9$ & $60.7$ & $81.0$ & $43.9$ & $50.4$ & $80.2$ & $60.5$ & $77.8$ & $82.2$ & $63.9$ \\
LPSAD \cite{9340837} [Our Implementation]
& $50.4$ & $57.7$ & $62.4$ & $79.4$ & $43.2$ & $53.2$ & $80.2$ & $57.5$ & $71.7$ & $78.5$ & $62.5$  \\
TORNADO-Net \cite{gerdzhev2020tornadonet} w/ Fusion
& $54.0$ & $59.8$ & $65.4$ & $80.9$ & $44.1$ & $53.9$ & $80.1$ & $63.9$ & $76.9$ & $81.8$ & $68.0$  \\
EfficientLPS \cite{sirohi2021efficientlps} 
& $59.2$ & $62.8$ & $70.7$ & $82.9$ & $51.8$ & $62.7$ & $80.6$ & $\textbf{71.5}$ & $\textbf{84.1}$ & $84.3$ & $69.4$  \\
\hline
 \textbf{\method} [\textcolor{blue}{Ours}] 
& $\textbf{61.0}$ & $\textbf{67.5}$ & $\textbf{72.0}$ & $\textbf{84.1}$ & $\textbf{56.0}$ & $\textbf{65.2}$ & $\textbf{85.3}$ & $66.0$ & $78.7$ & $82.9$ & $\textbf{75.8}$  
\\
\hline
\end{tabular}
}
\caption{Comparison of LiDAR panoptic segmentation performance on nuScenes\cite{caesar2020nuscenes} validation dataset.}
\label{bigtable_nu}}
\end{table*} 

\begin{table*}[htbp!]
{
\centering
\resizebox{1.82\columnwidth}{!}{
\begin{tabular}{l|cccc|ccc|ccc|cccccccccc}
\hline 
Method & PQ  
& PQ\textsuperscript{$\dagger$}
& RQ
& SQ 
& PQ\textsuperscript{Th} 
& RQ\textsuperscript{Th}
& SQ\textsuperscript{Th}  
& PQ\textsuperscript{St}  
& RQ\textsuperscript{St}  
& SQ\textsuperscript{St}  
&  mIoU  \\
\hline

LPSAD \cite{9340837} [Our Implementation]
& $22.5$ & $32.7$ & $34.0$ & $53.5$ & $18.7$ & $25.7$ & $70.5$ & $24.0$ & $37.1$ & $47.1$ & $35.5$  \\
TORNADO-Net \cite{gerdzhev2020tornadonet} w/ Fusion
& $33.7$ & $43.3$ & $46.0$ & $68.4$ & $41.2$ & $49.6$ & $83.1$ & $30.9$ & $44.7$ & $62.9$ & $44.5$  \\
DS-Net \cite{hong2020lidar}
& $35.6$ & $45.9$ & $49.2$ & $\textbf{68.6}$ & $27.4$ & $33.8$ & $76.8$ & $38.7$ & $55.0$ & $\textbf{65.5}$ & $54.5$  \\
\hline
 \textbf{\method} [\textcolor{blue}{Ours}] 
 & $\textbf{48.7}$ & $\textbf{60.3}$ & $\textbf{63.7}$ & $61.3$ & $\textbf{61.6}$ & $\textbf{71.7}$ & $\textbf{86.4}$ & $\textbf{43.8}$ & $\textbf{60.8}$ & $51.8$ & $\textbf{61.8}$ 
\\
\hline
\end{tabular}
}
\caption{Comparison of LiDAR panoptic segmentation performance on SemanticPOSS\cite{pan2020semanticposs} validation dataset.}
\label{bigtable_poss}}
\end{table*} 

\subsection{Quantitative evaluation}
\label{sec:quanti}
Starting from SemanticKITTI test results (the only available test benchmark), we can see that \method~ is standing out as the best framework for panoptic segmentation achieving PQ of $60.0\%$. Moreover, the method does not suffer from any degradation in semantic segmentation results, topping all other methods at  mIoU of $70.8\%$. Focusing on the methods that do not use object proposals, the new method, i.e., \method, performs outstandingly better than prior arts in terms of PQ such as LPSAD ($+22.0\%$), Panoster ($+7.3\%$) and DS-Net ($+4.1\%$). When we combine all methods, as it is shown in Table~\ref{bigtable}, \method~ can achieve better results than EfficientLPS \cite{sirohi2021efficientlps} by a small margin, but provides much better semantic segmentation results, and all without using any object proposal information which normally means access to bounding box information of dynamic objects.

To show the effectiveness of the proposed method in other large-scale outdoor datasets, and to show the generalization quality of \method, we also provide the validation benchmark  results for nuScenes and SemanticPOSS in Tables \ref{bigtable_nu} and \ref{bigtable_poss}, respectively. \method~ outperforms the prior-art in almost all the metrics for both datasets proving the proposed framework can work across a wide range of sensor setups and geometrical differences.  
\vspace{-2px}

\subsection{Qualitative evaluation}
\label{sec:quali}

\begin{figure*}[htbp] 
    \centering
    \includegraphics[width=\textwidth]{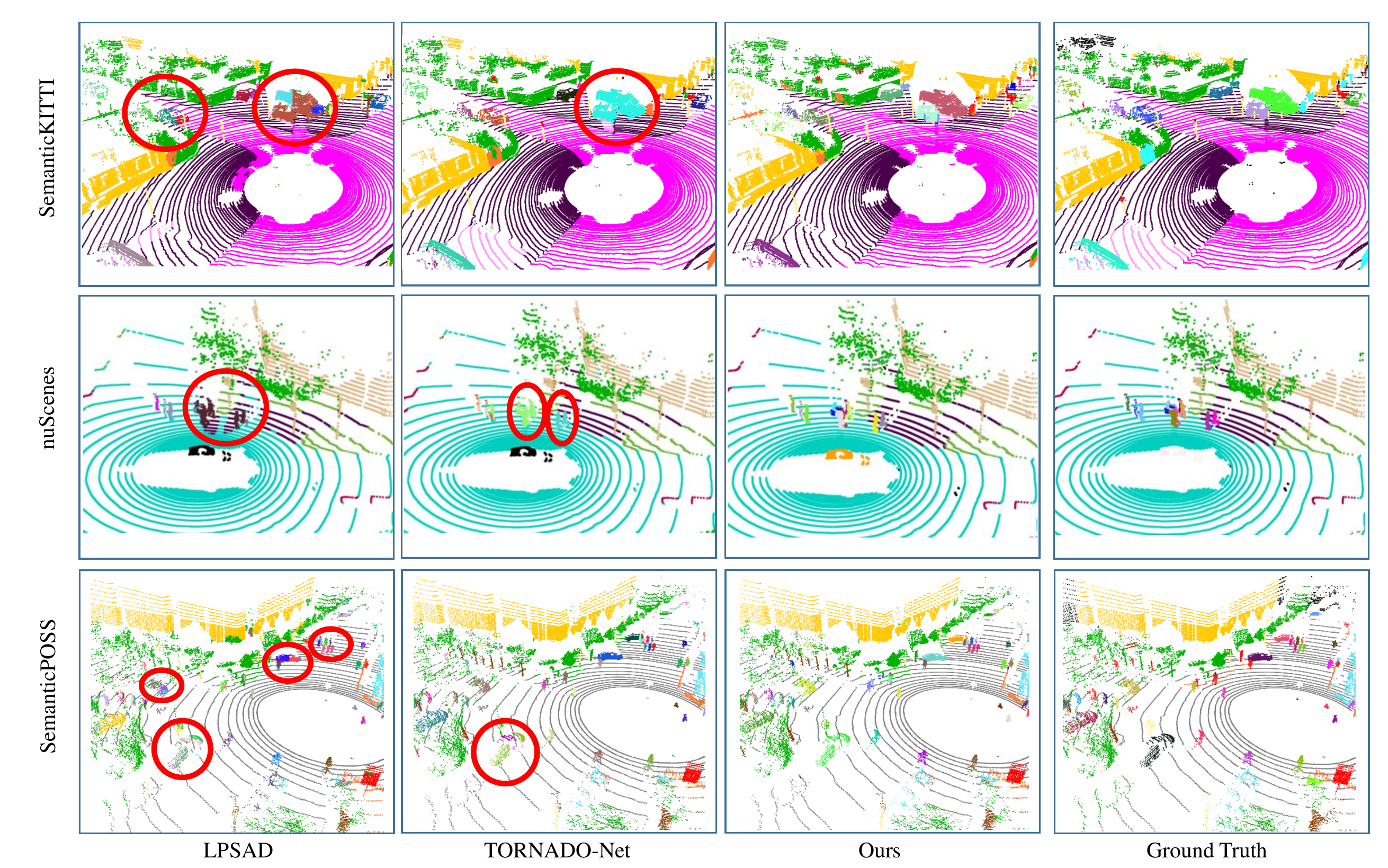}
    \caption{Comparison of  different models with \method~on three different datasets of SemanticKITTI, nuScenes, and SemanticPOSS. \textcolor{red}{Red} circles demonstrate that our method performs better in many details than recent state-of-the-art models.
    }
    \label{fig:Skitti_nuScenes_error}
\end{figure*}

As shown in Figure~\ref{fig:Skitti_nuScenes_error}, LPSAD \cite{9340837} and TORNADO-Net \cite{gerdzhev2020tornadonet}, which both are based on point offset predictions and clustering, perform poorly in crowded scenes where they have failed to differentiate instances of pedestrians and vehicles (visible in nuScenes and SemanticKITTI samples). In contrast, \method, which is a graph-based approach, is learned to separate most of the objects, regardless of their closeness, with much less confusion in assigning instance labels.

Another case is large objects like trucks as it is shown in the SemanticPOSS sample. It is clear that neither of the methods, i.e. LPSAD and TORNADO-Net, can solve the problem of less-segmentation for larger objects. Similar issues can be observed in SemanticKITTI as shown in Figure~\ref{fig:kyber_intro} on better benchmarks such as DS-Net~\cite{hong2020lidar}. However, \method~ can resolve this issue by assigning all those clusters into on instance, in this case truck, correctly.

\subsection{Ablation studies}
\label{sec:ablation}

To further demonstrate the effectiveness of our graph-based network, we provide an ablation analysis in Table \ref{tab:Ablation} comparing our baseline with our proposed method on validation sequence of SemanticKITTI dataset. The baselines are our semantic segmentation backbone followed by different clustering algorithms with no instance head. It can be observed that Meanshift \cite{comaniciu_meanshift}, performs relatively better than HDBSCAN \cite{campello_hdbscan}, and DBSCAN \cite{ester_dbscan} due to its robust kernel function insensitive to the density change in the point cloud. The proposed method outperforms all the baselines by a large margin.

\begin{table}[htb!]
\begin{center}
\scalebox{0.8} {
\begin{tabular}{ c|c|c|c|c  }
\hline 
\multicolumn{1}{c|}{\textbf{Architecture}} &
\multicolumn{1}{c|}{\textbf{\textbf{PQ}}} &
\multicolumn{1}{c|}{\textbf{\textbf{PQ}\textsuperscript{Th}}} &
\multicolumn{1}{c|}{\textbf{\textbf{RQ}\textsuperscript{Th}}} &
\multicolumn{1}{c}{\textbf{\textbf{SQ}\textsuperscript{Th}}} \\ 

 \hline \hline 
\multirow{1}{*}{Baseline w/ DBSCAN \cite{ester_dbscan}}
& $49.4$ & $40.5$ & $51.3$      &    $78.4$   \rule{0pt}{3ex}\\ \hline

\multirow{1}{*}{Baseline w/ HDBSCAN \cite{campello_hdbscan}}
& $53.4$ & $50.1$ & $63.1$      &    $78.5$   \rule{0pt}{3ex}\\ \hline

\multirow{1}{*}{Baseline w/ MeanShift \cite{comaniciu_meanshift} }
& $55.8$ & $55.7$ & $66.7$      &    $82.5$   \rule{0pt}{3ex}\\ \hline

\multirow{1}{*}{Proposed}
& $\textbf{63.3}$ & $\textbf{70.2}$ & $\textbf{80.1}$    &   $\textbf{86.2}$  \rule{0pt}{3ex}\\ \hline 

\end{tabular} 
} 
\end{center}
\caption{Ablation study of the proposed method vs baselines evaluated on SemanticKITTI \cite{DBLP:conf/iccv/BehleyGMQBSG19} validation dataset.}
\label{tab:Ablation}
\vspace{-14px}
\end{table}

\section{Conclusion}
\label{sec:conclusion}
In this work, we proposed GP-S3Net, a novel one-stage panoptic segmentation network. In addition to the strong semantic segmentation backbone, GP-S3Net has an effective instance network that takes in the clusters of the semantic masks of the \textit{thing} classes and constructs a graph with node and edge features learned from sparse convolutions. A GCNN is followed to predict whether an edge exists between each node pair. This novel approach has turned an unsupervised clustering task into a supervised graph edge classification problem where edges between two nodes indicate them belonging to the same instance. Comprehensive benchmarking results are presented on SemanticKITTI, nuScenes and SemanticPOSS datasets, demonstrating the state-of-the-art performance of \method.


{\small
\bibliographystyle{ieee_fullname}
\bibliography{egbib}
}

\end{document}